\documentclass{article}

\usepackage[preprint]{neurips_2025}

\usepackage[utf8]{inputenc} 
\usepackage[T1]{fontenc}    
\usepackage{hyperref}       
\usepackage{url}            
\usepackage{booktabs}       
\usepackage{amsfonts}       
\usepackage{nicefrac}       
\usepackage{microtype}      
\usepackage{xcolor}         

\usepackage{booktabs}
\usepackage{multirow,multicol}
\usepackage{wrapfig}

\usepackage{graphicx}

\usepackage{amsmath}

\usepackage{cleveref}

\usepackage{algorithm}
\usepackage{algpseudocode}

\usepackage{listings}
\usepackage{etoolbox}
\usepackage{longtable}

\usepackage{tcolorbox}
\tcbuselibrary{breakable}

\newcommand{\methodname}{\textsc{EDIT}}
\newcommand{\methodfullname}{\textbf{E}fficient \textbf{D}ynamic \textbf{I}nference \textbf{T}rimming}

\newlength{\ArrowNumSep}
\newcommand{\incl}[2]{#1\cellcolor{red!10}\textcolor{red}{\hspace{0.2em}($\uparrow$\hspace{\ArrowNumSep}#2)}}
\newcommand{\decl}[2]{#1\cellcolor{green!10}\textcolor{green!70!black}{\hspace{0.2em}($\downarrow$\hspace{\ArrowNumSep}#2})}
\newcommand{\inca}[2]{#1\cellcolor{green!10}\textcolor{green!70!black}{\hspace{0.2em}($\uparrow$\hspace{\ArrowNumSep} #2)}}
\newcommand{\deca}[2]{#1\cellcolor{red!10}\textcolor{red}{\hspace{0.2em}($\downarrow$\hspace{\ArrowNumSep}#2})}

\usepackage{amsmath}
\usepackage{amsfonts}
\usepackage{amssymb}
\usepackage{wrapfig}
\usepackage{subcaption}
\usepackage{multirow}
 \usepackage{mathtools} 

\usepackage{verbatim}

\usepackage{anyfontsize}

\usepackage{microtype}
\usepackage{graphicx}
\usepackage{booktabs} 
\usepackage{multirow}
\usepackage{amsmath,amssymb}
\usepackage{booktabs}
\usepackage{caption,subcaption}

\usepackage{xcolor}
\definecolor{mygreen}{HTML}{3cb44b}
\definecolor{skyblue}{HTML}{beffff}
\definecolor{lightgreen}{HTML}{90ee90}

\usepackage{color, colortbl}

\definecolor{emerald}{rgb}{0.31, 0.78, 0.37}

\usepackage{tcolorbox}
\usepackage{enumitem}
\setitemize{itemsep=10pt,topsep=0pt,parsep=0pt,partopsep=0pt}
\usepackage{colortbl}

\usepackage{xcolor}
\definecolor{mygreen}{HTML}{3cb44b}
\colorlet{myyellow}{green!10!orange!90!}
\makeatletter

\usepackage{tikz}
\usetikzlibrary{arrows,shapes,snakes,automata,backgrounds,fit,petri}
\usepackage{adjustbox}

\newcommand{\RN}[1]{%
	\textup{\lowercase\expandafter{\it \romannumeral#1}}%
}
\usepackage{tabu}

\newcommand{\beq}{\vspace{0mm}\begin{equation}}
\newcommand{\eeq}{\vspace{0mm}\end{equation}}
\newcommand{\beqs}{\vspace{0mm}\begin{eqnarray}}
\newcommand{\eeqs}{\vspace{0mm}\end{eqnarray}}
\newcommand{\barr}{\begin{array}}
\newcommand{\earr}{\end{array}}

\newcommand{\Rcal}{\mathcal{R}}

\usepackage{color, colortbl}
\definecolor{Gray}{gray}{0.93}

\usepackage{lipsum}

\usepackage{pifont}
\usepackage{makecell}

\usepackage{xcolor,amsmath}

\title{From Long to Short: LLMs Excel at Trimming Own Reasoning Chains}

\author{%
  Wei Han\thanks{Preprint. Work in Progress.} \\
  Independent Researcher \\
  \And
  Geng Zhan \\
  University of Sydney \\
  \And
  Sicheng Yu \\
  Singapore Management University \\
  \And
  Chenyu Wang \\
  University of Sydney \\
  \And
  Bryan Hooi \\
  National University of Singapore \\
}

\begin{document}

\maketitle
\begin{abstract}
O1/R1-style large reasoning models (LRMs) signal a substantial leap forward over conventional instruction-following LLMs. By applying test-time scaling to generate extended reasoning paths, they establish many SOTAs across a wide range of complex reasoning tasks. 
However, recent studies show that LRMs are prone to suffer from~\textit{overthinking}—the tendency to overcomplicate simple problems, leading to excessive strategy switching and long, convoluted reasoning traces that hinder their interpretability.
To mitigate this issue, we conduct a systematic investigation into the reasoning efficiency of a broad set of LRMs and uncover a common dilemma: the difficulty in balancing multiple generation objectives such as correctness and brevity. 
Based on this discovery, we propose a test-time scaling method, \methodname (\methodfullname), which efficiently guides LRMs to identify the shortest correct reasoning paths at test time.
\methodname~employs constraint-guided generation while jointly tracking length and answer distributions under varying constraints, allowing it to select responses that strike an optimal balance between conciseness and correctness. Extensive experiments across diverse models and datasets show that~\methodname~substantially enhance the reasoning efficiency, producing compact yet informative outputs that improve readability and user experience.
\end{abstract}
\section{Introduction}
The rapid emergence of large language models (LLMs) has revolutionized the development of artificial intelligence~\cite{xu2025towards}. 
These models possess the remarkable capability to autonomously complete a wide range of tasks across diverse scenarios following human instructions~\cite{grok2025grok3,yang2024qwen2,anil2023palm,grattafiori2024llama,touvron2023llama,achiam2023gpt}.
Recently, researchers have begun to push beyond traditional instruction-following or few-shot chain-of-thought. They explore more sophisticated application scenarios, where LLM has to trigger deliberate thinking and reasoning to produce the correct answer~\cite{fan2023nphardeval,wei2025browsecomp,yuan2025naturalreasoning}.
It has been observed that LLMs can naturally acquire advanced reasoning abilities through a specialized training procedure that combines supervised fine-tuning (SFT) as cold-start plus a subsequent reinforcement learning (RL) stage~\cite{guo2025deepseek}.
Building on this insight, an increasing number of large reasoning models (LRMs) have been introduced, marking key milestones in this area. 

However, the enhanced reasoning abilities of LRMs always come at the cost of increased output redundancy~\cite{chen2024not}, leading to substantial computational overhead that grows quadratically with output length. 
In addition, the excessively long reasoning chains generated by these models make it difficult for users to interpret and quickly grasp the underlying logic during interactions.
Recent models like Qwen-3~\cite{qwenteam2025qwen3}~and Kimi~\cite{team2025kimi,team2025kimik2}~combine mixed thinking/non-thinking data or length penalty into training, but still fail to make LRMs wisely select the mode during the inference time. 

To alleviate this issue, we propose a test-time scaling method---\methodname~(\methodfullname), which leverages the variation trends of solution lengths and answer confidence under parametric constraints. 
Building on this statistical discovery, we design a search algorithm that  automatically identifies the critical point at which LRMs achieve the \textit{pareto optimality}~between reasoning cost and accuracy. 

In summary, our main contributions are as follows:
\begin{itemize}
    \item We explore the innovative task named reasoning path compression, with a particular focus on constraint-based generation. Our study includes a systematic analysis on the characteristics of the independent and joint distribution of reasoning length and correctness under various restrictions fed to the LLMs.
    \item Based on the previous analysis, we propose~\methodname~(\methodfullname), a fully test-time reasoning path pruning framework. 
    At the core of this framework is a~\textit{dual-goal search}~algorithm that effectively guides the model toward near-optimal solutions in terms of both brevity and correctness. 
    To address the limitations of conventional length measurements, we introduce two evaluation approaches---one metric and one comparison paradigm that better aligns with the observed trade-offs between length and correctness.
    \item Extensive experiments on seven advanced reasoning and instruction-following models as well as three typical datasets validate the effectiveness of \methodname, showcasing its ability to generate concise and accurate reasoning paths.
\end{itemize}
\section{Related Work}
\paragraph{Large Reasoning Models.} Reasoning is a desirable capability for LLMs. The first trial to elicit LLM's reasoning ability was on mathematical problems~\cite{hendrycks2measuring}, where outcome- or process-based verifiers are trained to supervise LLM learning~\cite{lightman2023let}. More recently, researchers believe that LLMs can simulate human's System 1 (fast, instinct) and System 2 (slow, delibrate)~\cite{yu2024distilling} to efficiently handle logical tasks. 
Reward-guided decoding~\cite{khanovargs,liao2025reward,xie2023self} and Monte Carlo tree search (MCTS)~\cite{hao2023reasoning,zhang2024rest} improve these base methods by introducing external or future feedback.
CoT~\cite{wei2022chain}~and several subsequent enhancement work~\cite{wang2022self,yao2023tree,besta2024graph}~
managed to elicit the reasoning paths directed to the correct answer.
The release of OpenAI-o1~\cite{jaech2024openai} heightened the engagement from the research community in test-time scaling. 
DeepSeek-R1~\cite{guo2025deepseek}~demonstrates the great role of reinforcement learning in LLM's grasping of reasoning capabilities~\cite{liu2025understanding}, and there are a line of follow-up works to develop skilled reasoning models~\cite{shen2025vlm,liu2025visual}. Regards foundation LRM training, most efforts are put at the RL algorithm~\cite{yu2025dapo}~and self-evolving pipelines~\cite{zelikman2022star,shao2024deepseekmath}.

\paragraph{Underthinking and Overthinking Issue in LRMs.} While O1/R1-like LRMs exhibit stunning capabilities in solving complex problems, they expose undesired weakness termed~\textit{overthinking}~when encountering simple questions like ``\textit{what is the result of 2+3?}''~\cite{chen2024not,team2025kimi,chiang2024over}. Meanwhile, although LRMs are encouraged to generate complete reasoning paths based on the SFT versions, 
\textit{underthinking}~\cite{wang2025thoughts}~is ubiquitous among LRMs, which could stagate at shallow thinking. 
To alleviate this issue, the model should learn to avoid  both~\textit{underthinking} (incorrection)~and~\textit{overthinking}~(inefficiency). 
Several routines have been proposed to alleviate reasoning inefficiency while maintaining an acceptable error rate: direct tuning on pruned generation~\cite{xia2025tokenskip,han2024token,ma2025reasoning}, latent-space reasoning~\cite{geiping2025scaling,hao2024training}~and input-aware test-time searching~\cite{damani2024learning,wang2025sampling}. While Most of these approaches focus on system-level improvement, such as reducing memory usage or latency, this work falls under the category of test-time scaling. However, unlike prior trials,~\methodname~aims at generating concise and interpretable solution.
\section{Methods}
In this section, we first formalize the task definition of efficient reasoning mathematically. 
Next, we present key observations derived from our benchmark results. 
Finally, based on these observations, we propose our efficient search algorithm that navigates toward the optimal solution.

\subsection{Problem Formulation}
\methodname~is a test-time adaptation method, and therefore we hold a target model $\mathcal{M}$~with constant parameters during the entire inference period. 
Given a reasoning dataset $\mathcal{D}$, for each question $q_i\in\mathcal{D}$ with the ground truth $a_i$, 
we prompt $\mathcal{M}$~multiple times to sample a collection $\Rcal =\{\mathbf{r}_1,\mathbf{r}_2,...,\mathbf{r}_N\}$, where
The final answer $\hat{a}_i=\phi(\mathbf{r}_i)$ can be further extracted  by a format matching function $\phi$. Our goal is to maximize the accuracy, which is based on an identity-checking function: 
\begin{equation}
    \mathrm{acc}=\frac{\sum_{\mathbf{r}_i\in \mathcal{R}}\mathbf{1}(\phi(\mathbf{r}_i)=a_i)}{|D|}
\end{equation}
where $\mathbf{1(\cdot)}$ is the indicator function and $|D|$ is the dataset size. Meanwhile, we aim to minimize the average reasoning lengths $\Bar{l}$ calculated as the average number of tokens from the correct solution path for each question (at most one for each question) across the datasets. 
\begin{equation}
    \Bar{l} = \frac{\sum_{i,\phi(\vec{r}_i=a_i)} l_i}{|D_{corr}|}
\end{equation}
The goal is to seek a generation strategy $\Gamma$, which can produce accurate answers and minimize the length of reasoning paths. 
Formally, this problem can be rewritten in a constrained optimization form, where $\alpha$ is the length constraint.
\begin{equation}
    \Gamma^* = \arg\max_\Gamma \mathrm{acc}(\Gamma), s.t.\; \Bar{l} = \Bar{l}(\Gamma)\le \alpha 
\end{equation}
Empirically, as $\alpha$ diminishes, accuracy usually degrades because more solutions are truncated ahead of figuring out the answer. However, we can leverage this property for efficiency evaluation.

\subsection{Statistical Findings from Constrained Reasoning}
We are curious about the generation distributions after adding length constraints to the prompts. 
Since on a generic application scenario with deep thinking enabled, no explicit limitations on generation lengths are applied---LRMs are granted the full freedom to extend their reasoning chains. 
To dive deeper, we curate a simple parametric prompt template as illustrated in~\Cref{tab:constraint_prompt}. We benchmark the responses by sampling multiple responses per question on the MATH500 dataset and under different constraints.

\label{sec:findings_constraint}
\begin{figure}[ht]
    \centering
    \begin{subfigure}[t]{0.35\textwidth}
        \includegraphics[width=\linewidth, trim=1cm 0 0.25cm 0]{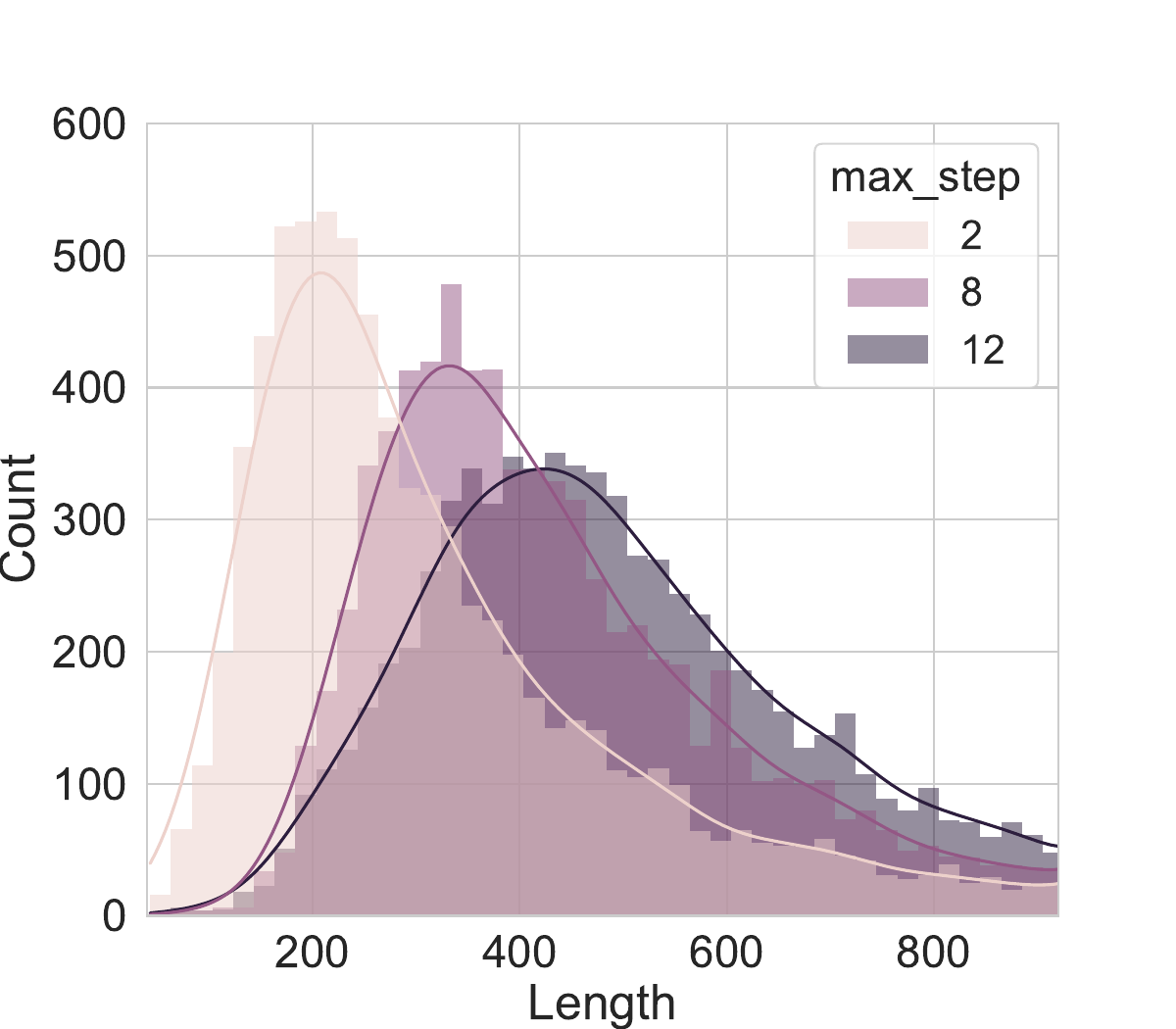}
        \caption{Length distribution under three constraints parameterized by the maximum steps that can be taken to solve each problem.}
        \label{fig:length_dist}
    \end{subfigure}
    \hspace{0.01\linewidth}
    \begin{subfigure}[t]{0.61\textwidth}
        \includegraphics[width=\linewidth, trim=0.1cm 0 0 0]{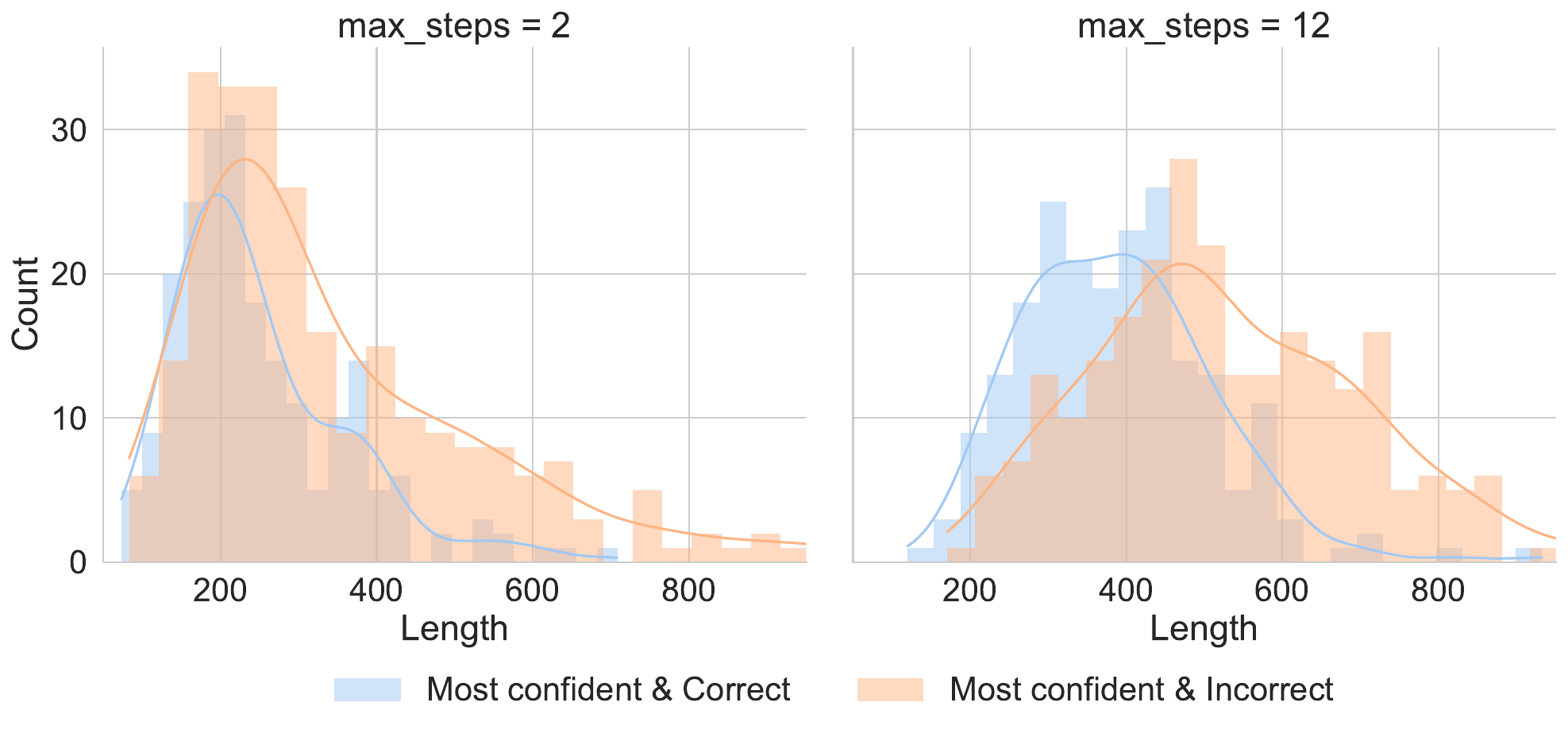}
        \caption{Length distributions of ``most confident and correct'' and ``most confident and wrong'' answers under two constraints. The distribution is separable when constraint is loose (left). Serious distribution overlap exhibits when applying stronger constraint.}
        \label{fig:distrib_binary}
    \end{subfigure}
    \caption{Histogram of length distributions on MATH500 dataset with LLaMA-3.1-8B: the generation length distributions for (a) all generated solutions, 16 samples per question (b) most confident answers, with binary justifications as correct/wrong. ``Count" is the total number of outputs that fall in the corresponding length interval. We sample 16 outputs per question.}
    \label{fig:length_benchmark}
\end{figure}

The length distributions of outputs using this template in MATH500 test set and LLaMA-3.1-8B model are plotted in~\Cref{fig:length_dist}.  
We highlight the most confident answer for each problem as well as the corresponding correctness in~\Cref{fig:distrib_binary}. 
Generally, the length distribution is long-tailed, especially when the constraint is stringent.
Besides, there are two notable observations from the plots: 
\begin{itemize}
    \item In~\Cref{fig:length_dist}, LLM is sensitive to the imposed constraints and can adapt its output distribution accordingly---the overall length distribution shifts right (i.e., longer) as the constraint loosens (i.e., more steps allowed).
    \item In~\Cref{fig:distrib_binary}, as the constraint becomes more straingent (right to left), incorrect answers (orange bars) is prone to dominate the shorter part of answers, i.e. the model  tends to prioritize the length and ignores correctness. 
\end{itemize}
These findings inspire us that we may start from free generation and use its generation as a signal to probe potential better solutions in tighter constraints. In the following subsection, we are going to introduce the core algorithm,~\methodname, in detail.

\subsection{Dual Goal Search}
\begin{figure}[ht]
    \centering
    \begin{subfigure}[b]{0.49\textwidth}
        \includegraphics[page=1, width=\linewidth, trim=0 0 0 0]{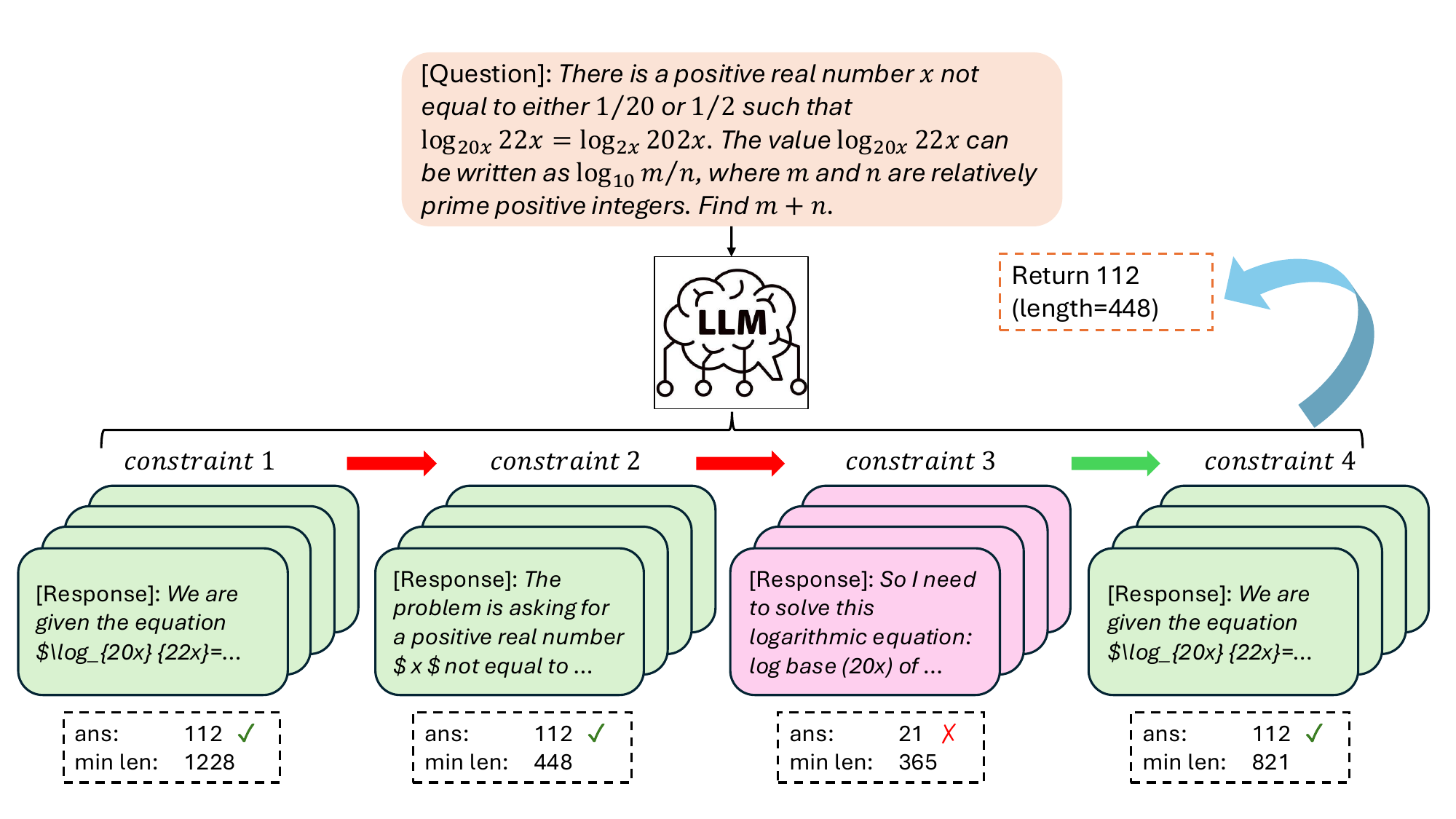}
        \caption{Cross-constraint checking fails ($\beta_0=0$).}
        \label{fig:edit_example1}
    \end{subfigure}
    \begin{subfigure}[b]{0.49\textwidth}
        \includegraphics[page=2, width=\linewidth, trim=0 0 0 0]{figs/method.pdf}
        \caption{Cross-constraint checking succeeds ($\beta_0=1$).}
        \label{fig:edit_example2}
    \end{subfigure}
    \label{fig:edit_example}
    \caption{Two running examples of the~\methodname~framework, with and without patience respectively. Dark red and green arrows between search steps represent the constraint is tighter or relaxed.
    In figure (a), the consistency checking fails due to running-out of patience. The next step continues with a looser constraint and a sub-optimal response is returned. A higher value of initial patience in figure (b) and a better response is returned.}
\end{figure}

\paragraph{Overview}
\methodname~is inspired by the achievements of \textit{principle-guided generation}~\citep{sun2023principle,pmlr-v235-beurer-kellner24a,pmlr-v202-zhou23g}~on LLMs and LRMs' ultimate goals of response on queries~\cite{wu2025more}. In~\methodname, beyond the ultimate goal of solving the given problem, an additional parameterizable length constraint is included in the prompt templates. 
In normal scenarios, LRMs endeavor to satisfy both goals. However, when LRMs forecast that they are incapable of fulfilling both goals, they tend to sacrifice one objective in favor of the other. 
Take~\Cref{fig:edit_example1}~as an example---when the constraint is loose, LRMs automatically enhance the reasoning outcome by creating a long reasoning path, as developers expect. 
At the beginning of tightening the length constraint, LRMs successfully follow our guidance to shorten their output length while keeping the final answer correct (step 1 to step 2). However, when we further limit the output length, the LRMs cannot simultaneously satisfy both objectives forever. 
They tend to prioritize the goal they ---typically the length during generation, and output the wrong answer (from step 2 to step 3). 
To address this, \methodname~monitors the dynamics of two key statistics—answer confidence and length distribution—and adjusts the constraint strength accordingly, either relaxing or tightening it as needed.

\paragraph{Algorithm} We describe the core algorithm in~\methodname, \textit{Dual-Goal Search}~in~\Cref{alg:dual_search}. 
This algorithm is essentially a binary search with patience. 
The search process begins with the loosest constraint and full patience (line 3). 
At the start of every following iteration, the algorithm checks the consistency of the predicted answers and length distributions by comparing the answer and length metrics. Consistency is not broken only if 1) Answers from the two iterations are same; 2) The length metric adapts in response to changes in the applied constraints.

If the consistency checking passes,  the algorithm proceeds with a stronger constraint by updating the current constraint parameter as the new upper bound (line 9-10). 
Otherwise, if there is patience remaining, the algorithm continues this procedure with a tighter constraint and reduced patience (line 19-20). If patience is exhausted (line 11-18), the algorithm compares the current answer with the most confident answer recorded in previous iterations from $H$. The search boundary is updated according to the comparison result: if the current answer is considered consistent with the previously confident answer, the upper bound is decreased to further strengthen the constraint; otherwise, the lower bound is increased to make room for LRMs to think more freely and find the correct answer. 

The hyperparameter patience is introduced to prevent the misdirect of search and the return of suboptimal solutions. 
For example, as illustrated in~\Cref{fig:edit_example}, when patience $\beta=0$, cases like~\Cref{fig:edit_example1}~may arise, where the algorithm encounters an inconsistency and exits the process immediately. With the patience $\beta_0$, it can continue to explore with more aggressive constraints to ignore accidental errors.

The criteria of confidence can be implemented as majority voting or aggregation of logarithm probabilities. We find the results of the two implementations are quite close, and thus opt to use voting for simplicity. As for the statistical indicators that represent the length distribution, we select the average value of the minimum, lower quartile, and median, based on the findings from~\Cref{sec:findings_constraint}:
\begin{equation}
    lstat = \mathtt{Ans\_Stat}(\mathcal{C}, \hat{a}) = \frac{\min(\mathcal{L})+Q_1(\mathcal{L})+\rm{median}(\mathcal{L})}{3}
\end{equation}
where $\mathcal{L}$ denotes the length collection of responses whose predicted answers equal the most confident answer $\hat{a}$.

\begin{table}[ht]
    \label{tab:constraint_prompt}
    \caption{The prompt template to apply constraints and elicit principle-guided responses. Variables with curly braces surrounded are placeholders to be filled. ``num\_step" is the upper limit of reasoning steps we assign to LRMs.}
    \begin{tcolorbox}[colback=gray!5!white]
        [System] \textit{<The system prompt is omitted>}. 
        \newline
        
        [User] Solve the given math problem step by step. You must output the final answer in a box "\verb|\|\verb|\|\{\{final\_answer\}\}". You are limited to at most \{num\_step\} reasoning steps. Stop generating immediately and output the answer if you reach the maximum \{num\_step\} steps or obtained the final answer early.
        \\ \\
        Question: \{question\} \\ \\
        Solution:
    \end{tcolorbox}
\end{table}

\begin{algorithm}[t]
\caption{Dual-Goal Search}
\label{alg:dual_search}
\begin{algorithmic}[1]
\small
\State \textbf{Input:} Model $\mathcal{M}_\theta$, Question $q$, Maximum Number of Iterations $T$, Total Sampling Budget~$N$, Constraint Template $\Psi$ with Parameter $\tau=\tau_{max}$ (smaller $\tau$ means tighter constraint), patience $\beta=\beta_0$.
\State \textbf{Output:} Final Answer $\hat{a}$ \\

\textbf{Initialize} Number of Samples in each iteration $n = \frac{N}{T}$, History Record Memory $H$

\For{$t = 1$ to $T$}
    \State Sample candidates with parameterized constraints $\mathcal{C}_t = \{c_{t,1}, \dots, c_{t,n}\} \sim \mathcal{M}_\theta(q, \Psi(\tau_t))$
    \State $answer\_conf = \texttt{Cal\_Confidence}(\mathcal{C}_t)$ 
    \State $\hat{a}_t = \arg\max(answer\_conf)$
    \State $lstat_t = \texttt{Ans\_Stat}(\mathcal{C}_t, \hat{a}_t)$    \Comment{obtain the statistics of length distribution}
    \If{\texttt{Check\_Consistency}$(H, \hat{a}_t, lstat_t)$}
        \State $\tau_{max}=\tau_t$, $\tau_t = (\tau_{max} + \tau_{min}) / 2$
    \ElsIf {$\beta == 0$}    \Comment{checking consistency failed, patience spent up}
        \If{$\hat{a}_i==\texttt{most\_confident}(H)$}
            \State $\tau_{max} = \tau_t$
        \Else
            \State $\tau_{min} = \tau_t$
            \State continue
        \EndIf
        \State $\beta = \beta_0$
    \Else   \Comment{patience does not run out, continue to searching in shorter generations.}
        \State $\tau_t=(\tau_{t-1}+\tau_{min})/2$, $\beta = \beta - 1$
    \EndIf 
    \State $H = H \cup \{(\hat{a}_i, lstat_i)\}$
\EndFor
\State $\hat{a} = \texttt{most\_confident}(H)$
\State \Return $\hat{a}$
\end{algorithmic}
\end{algorithm}
\section{Experiments}
\label{sec:exp}
\subsection{Inference Setup}
\paragraph{Datasets} We focus on mathematical reasoning tasks. 
Three publicly available datasets are evaluated in this stage: GSM8K~\cite{cobbe2021training}, MATH500~\cite{lightman2023let}~and AIMO~\cite{patel2024aime}, the super collection of AIME 2022-2024. Specifications of these datasets can be found in~\Cref{app:ds_stats}.

\paragraph{Models} We assess~\methodname~on seven LLMs, covering various parameter sizes and training patterns. 
The targeted models include: non-reasoning instruction-following models (LLaMA-3.1-8B~\cite{grattafiori2024llama}), reinforcement learning enhanced reasoning models (QwQ-32B~\cite{qwenteam2024qwq}, Qwen3-8B/32B~\cite{qwenteam2025qwen3}~and~Phi-4-Reasoning~\cite{abdin2025phi}), LRM-distilled models~\cite{guo2025deepseek}~(R1-distilled-Qwen1.5B, R1-distilled LLaMA3-8B and Phi-4-reasoning-14B~\cite{abdin2025phi}). More specifications about targeted models can be found in~\Cref{app_model_spec}.

\paragraph{Generation Configuration} 
We keep the same set of sampling parameters, such as temperatures and top\_\textit{p} values across all baselines and our implementations to ensure a fair comparison. 
To ensure the fair comparison, the number of samples $N$ is kept the same in all experiments. As a result, for methods using iterative algorithms, the number of sample per iteration is upper bounded.

\subsection{Baselines and Metrics}
\paragraph{Baselines} We compare our methods with a series of advanced baselines:  Direct CoT Prompting (\textbf{DP})~\citep{kojima2022large}~forces the model to generate a step-wise thinking path. 
For Best-of-N (\textbf{BoN})~\citep{sunfast}, we gather a collection of solutions that obtain the most common answer according to a voting mechanism as in self-consistency~\citep{wang2022self}. We then pick the shortest solution from that collection as the final generation result. 
In Self-Truncation~(\textbf{ST}), similar to the implementation in \cite{xia2025tokenskip}, we first prompt LRMs to generate reasoning chains and select the chains that produce the most common answers. Then we continue to prompt LRMs to prune these solution paths. All baselines are running under a fixed budget $N=64$ samples (except direct prompting, where LRM generates solutions using greedy decoding) no matter how many iterations each method has gone through.

\paragraph{Metrics} The basic metrics are accuracy and average length, both of which are averaged over the whole datasets. However, for the length comparison, we find treating correct and incorrect responses equally causes inconsistency on the intuitive expression of the models' overall performance. Take Llama-3.1-8B on MATH500 test set as an example, the DP has an average response length of 110 tokens, while~\methodname~spends more tokens (121) and acquires twice the accuracy (61.7\% v.s. 30.2\%). The measurement of length becomes meaningless at this time. Hence, a calibration approach is required—one that considers both accuracy and the cost associated with incorrect predictions. 
We are going to introduce the calibrated metric in~\Cref{sec:efficiency_measure}, and show that with this penalty item, the newly calibrated lengths are more meaningful to reflect LRMs' capability.

\begin{table}[ht]
    \centering
    \small
    \caption{Result on three math problem datasets. The \textbf{best accuracy}~among all baseline methods are~\underline{underlined}. Accuracy and relative length variation with the best baselines are highlighted in the brackets.
    }
    \resizebox{\linewidth}{!}{
        \begin{tabular}{l|l|*{2}{ccc}cc|cc}    
        \toprule
          \multirow{2}{*}{Dataset} & \multirow{2}{*}{Model}  & \multicolumn{2}{c}{DP} && \multicolumn{2}{c}{BoN} && \multicolumn{2}{c|}{ST} & \multicolumn{2}{c}{\methodname}  \\
           ~ & ~  & Acc$\uparrow$ & Length$\downarrow$ && Acc$\uparrow$ & Length$\downarrow$ && Acc$\uparrow$ & Length$\downarrow$ & Acc$\uparrow$ & Length$\downarrow$ \\
           \midrule
          \multirow{7}{*}{GSM8K} & Qwen3-32B  & 88.2 & 725 && \underline{94.2} & 181 && 94.0 & 195 & \inca{94.4}{0.2} & \decl{83}{54.1\%} \\
           ~ & Llama-3.1-8B & 67.2 & 302 && \underline{72.3} & 120 && 68.4 & 84 & \deca{71.4}{0.9} & \decl{87}{27.5\%}  \\
           ~ & Qwen3-8B & 88.2 & 380 && \underline{93.4} & 202 && 93.3 & 163 & 93.4 & \decl{103}{36.8\%} \\
           ~ & QwQ-32B & 88.0 & 706 && \underline{93.0} & 378 && 92.6 & 765 & \inca{93.3}{0.3} & \decl{223}{41.0\%} \\
           ~ & Phi-4-Reasoning & 82.6 & 524 && 91.1 & 419 && \underline{91.5} & 370 & \inca{91.8}{0.3} & \decl{286}{22.7\%} \\
           ~ & RD-Qwen-1.5B & 74.1 & 476 && \underline{86.4} & 290 && 85.9 & 236 & \deca{85.7}{0.7} & \decl{154}{46.9\%} \\
           ~ & RD-Llama-3.1-8B & 65.2 & 420 && \underline{89.1} & 330 && 88.2 & 295 & \deca{88.7}{0.4} & \decl{262}{20.6\%} \\
           \midrule
          \multirow{7}{*}{MATH} & Qwen3-32B & 52.0 & 480 && \underline{71.6} & 328 && 68.8 & 368 & \deca{70.8}{0.8} & \decl{297}{9.4\%} \\
           ~ & Llama-3.1-8B & 30.2 & 110 && \underline{52.6} & 131 && 51.0 & 98 & \deca{52.4}{0.2} & \decl{108}{17.6\%}  \\
           ~ & Qwen3-8B & 59.8 & 480 && 73.4 & 353 && \underline{74.2} & 291 & \deca{73.2}{1.0} & \decl{207}{28.9\%} \\
           ~ & QwQ-32B & 43.8 & 612 && \underline{64.4} & 394 && 61.0 & 327 & \inca{64.8}{0.4} & \decl{298}{24.4\%} \\
          ~ & Phi-4-Reasoning & 73.5 & 1033 && \underline{81.2} & 796 && 78.6 & 535 & \deca{80.2}{1.0} & \decl{335}{57.9\%} \\
           ~ & RD-Qwen-1.5B & 48.4 & 520 && 60.4 & 396 && \underline{62.6} & 284 &  \deca{62.4}{0.2} & \decl{249}{12.3\%} \\
           ~ & RD-Llama-3.1-8B & 51.4 & 406 && 62.4 & 411 && \underline{63.0} & 374 & \inca{65.0}{2.0} & \decl{337}{9.9\%}  \\
           \midrule

          \multirow{7}{*}{AIMO} & QwQ-32B & 72.2 & 7352 && 84.4 & 5480 && \underline{86.7} & 4980 & 86.7 & \decl{3791}{30.8\%}  \\
           ~ & Llama-3.1-8B & 5.6 & 13 && 7.8 & 47 && \underline{7.8} & 32 & \inca{11.1}{3.3} & \incl{40}{25.0\%}  \\
           ~ & Qwen3-8B & 58.9 & 3335 && \underline{76.7} & 3187 && 75.6 & 3045 & \deca{74.4}{2.3} & \decl{2284}{28.3\%} \\
           ~ & Qwen3-32B & 77.8 & 6189 && 82.2 & 5533 && \underline{83.3} & 4980 & 83.3 & \decl{4140}{16.9\%} \\
           ~ & Phi-4-reasoning & 68.9 & 3565 && \underline{82.2} & 3870 && 77.8 & 2921 & \inca{83.3}{1.1} & \decl{2533}{34.5\%} \\
           ~ & RD-Qwen-1.5B & 22.9 & 2843 && \underline{42.2} & 1728 && 36.7 & 1155 & \deca{41.1}{1.1} & \decl{1376}{20.4\%} \\
           ~ & RD-Llama-3.1-8B & 51.1 & 2578 && \underline{60.0} & 2348 && \underline{60.0} & 2149 & 60.0 & \decl{1770}{17.6\%} \\
           \bottomrule
        \end{tabular}
    }
    \label{tab:res_3math}
\end{table}

\subsection{Main Results}
\label{sec:main_results}
The results averaged over five runs are shown in~\Cref{tab:res_3math}. The key findings from these results are following:
\begin{itemize}
    \item RL-enhanced models (e.g., the Qwen-3 8B/32B, QwQ-32B and Phi-4-Reasoning) consistently show better capabilities than models in other categories. Models trained with reasoning data distillation (e.g., R1-Distill-Llama-3.1-8B and R1-Distill-Qwen1.5B) perform worse than RL-enhanced models but still outperform the SFT-only baseline (Llama-3.1-8B).
    \item Direct CoT prompting (DP) baseline always ranks at the bottom due to the lack of test-time compute scaling. On the other hand, in several test cases, DP can produce the shortest reasoning path compared to other methods, which could be attributed to the impact of greedy decoding. By scaling test-time compute, best-of-N (BoN) selection and self-truncation (ST)~both achieve significantly better results compared to DP. 
    \item While showing the highest accuracy in every test cases, \methodname~shows superior reasoning efficiency---it manages to preserve accuracy to the maximum extent and cut off the reasoning path. In most cases, \methodname~nearly achieves or surpass the best baseline with concise reasoning process. Specifically, the largest performance gap is below relatively 2\% with 20\% fewer tokens produced in the reasoning paths.
    \item \methodname~appears to struggle for decision on non-reasoning models. The only outlier on length is on AIMO using Llama-3.1-8B, where~\methodname~generate 25\% more tokens to pursue a significant accuracy boost. In contrast, on MATH and GSM8K, it works similar to other models that strive to reach the pareto optimality.
    
\end{itemize}


\subsection{Efficiency Measure}
\label{sec:efficiency_measure}
In~\Cref{sec:main_results}, evaluations on two major metrics were reported. However, these two metrics can conflict with each other, obscuring a clear comparison between methods. Admittedly, efficiency (reasoning path length) is important, but we must remain aware that it is sometimes acceptable for the model to sacrifice a degree of efficiency in exchange for improved accuracy. Hence, in evaluating efficiency, we should also quantify how much cost the model has spent on generating incorrect answers. For this purpose, we provide two evaluation methods as follows.

\subsubsection{Token-wise Efficiency}
\paragraph{Accuracy per Effective Length (AEU)}
Previous work~\citep{ma2025cot} adopts Accuracy per Computation Unit (ACU) as the efficiency indicator, which takes into account both the number of trainable parameters and the number of tokens. 
Since there are no trainable parameters in our experimental settings, 
we slightly modify this idea by removing the item of parameters and recalibrating the token counting methods, which can better reflect the integrated computational cost. 
Formally, we define $\mathrm{AEU}=\frac{ACC}{\bar{l}}$. 

\paragraph{Length Recalibration}
Recall the batched inference process in vLLM~\cite{kwon2023efficient}, where the time complexity is determined by the longest generated sequence. 
Hence, we formulate the effective length $\bar{l}_{pen}$ as
\begin{equation}
    \bar{l}_{pen} = \frac{\sum_i \mathbf{1}(\phi(\mathbf{r}_i)=a_i)l_{correct} +  \gamma (1-\mathbf{1}(\phi(\mathbf{r}_i)=a_i)) \max\{l_{wrong}\}}{|D|}
\end{equation}
where $\gamma>1$ is the penalty factor to adjust the weight of extra length cost. 
 
We report the re-calibrated length as well as AEU, as shown in~\Cref{tab:length_recali}.
There are many interesting findings from the table: 1) AEU score varies apparently across these datasets, which can serve as a difficulty indicator. Intuitively, if a question is more hard, the model has to take longer thinking process to reach the final answer. GSM8K exhibits the highest AEU value, and its difficulty is relatively low, correspondingly. In contrast, AIMO who owns the most challenging tasks, shows the lowest AEU.

\begin{table}[t]
    \centering
    \small
    \caption{Peformance under recalibrated length $\Bar{l}_{pen}$ and AEU. The best and second best results are \textbf{bolded}~and~with~\underline{underline}. AEU scores are rescaled by $10^2$ for better readability.}
    \begin{tabular}{l|l|cccccc|cc}
    \toprule
        \multirow{2}{*}{Dataset} & \multirow{2}{*}{Model} & \multicolumn{2}{c}{DP} & \multicolumn{2}{c}{BoN} & \multicolumn{2}{c|}{ST} & \multicolumn{2}{c}{EDIT} \\
        ~ & ~ & $\Bar{l}_{pen}\downarrow$ & AEU$\uparrow$ & $\Bar{l}_{pen}\downarrow$ & AEU$\uparrow$ & $\Bar{l}_{pen}\downarrow$ & AEU$\uparrow$  & $\Bar{l}_{pen}\downarrow$ & AEU$\uparrow$  \\ 
        \midrule
        \multirow{6}{*}{GSM8K} & Qwen3-32B  & 1154 & 7.64 & \underline{536} & \underline{17.57} & 614 & 15.30 & \textbf{217} & \textbf{43.50}  \\
        ~ & Llama-3.1-8B & 2543 & 2.64 & 559 & \underline{12.93} & \underline{532} & 12.85 & \textbf{469} & \textbf{15.22} \\
        ~ & Qwen3-8B & 824 & 10.70 & 638 & 15.04  & \underline{613} & \underline{15.22}  & \textbf{523} & \textbf{17.86} \\
        ~ & QwQ-32B & 1150 & 7.65 & \underline{621} & \underline{14.98} & 920 & 10.07 & \textbf{573} & \textbf{16.28} \\
        ~ & Phi-4-Reasoning & 1594 & 5.18 & \underline{862} & \underline{10.57} & 1215  & 7.49 & \textbf{821} & \textbf{11.18}  \\
        ~ & RD-Qwen-1.5B & 2195 & 3.38 & \textbf{1052} & \textbf{8.21}  & 1095 & 7.84 & \underline{1065} & \underline{8.05} \\
        ~ & RD-Llama-3.1-8B & 1782 & 3.66 & \underline{790} & \underline{11.28} & 1007 & 8.76 & \textbf{687} & \textbf{12.91} \\
        
        \midrule
        \multirow{7}{*}{MATH} & Qwen3-32B & 2083 & 2.50  & 1340 & 5.34 & \underline{1027} & \underline{6.70} & \textbf{763} & \textbf{9.28} \\
        ~ & Llama-3.1-8B & 2283 & 1.32 & 1028 & 5.12 & \underline{970} & \underline{5.20}  & \textbf{824} &  \textbf{6.36} \\
        ~ & Qwen3-8B & 1343 & 4.45 & 907 & 8.09 & \underline{844} & \underline{8.79} & \textbf{737} & \textbf{9.93} \\
        ~ & QwQ-32B & 2880 & 1.52 & 1563 & 4.12  & \underline{1159} & \underline{5.26} & \textbf{1139} & \textbf{5.69}   \\
        ~ & Phi-4-Reasoning & 1935 & 3.80 & 1607 & 5.05 & \underline{1518} & \underline{5.18} & \textbf{1497} & \textbf{5.40} \\
        ~ & RD-Qwen-1.5B & 1775 & 2.73 & 1242 & 4.86 & \underline{1100} & \underline{5.69} & \textbf{879} & \textbf{7.10} \\
        ~ & RD-Llama-3.1-8B & 1699 & 3.08 & 1232 & 5.06 & \underline{1200} & \underline{5.25} & \textbf{1046} & \textbf{6.21} \\
        \midrule
        \multirow{7}{*}{AIMO} & Qwen3-32B & 13408 & 0.54 & 10844 & 0.78 & \underline{9615} & \underline{0.90} & \textbf{8878} & \textbf{0.98} \\
        ~ & Llama-3.1-8B & 11623 & 0.05 & \textbf{11307} & 0.07 & \underline{11513} & \underline{0.07} & 12290 & \textbf{0.09} \\
        ~ & Qwen3-8B & 13397 & 0.44 & 10776 & 0.71 & \underline{9857} & \underline{0.77} & \textbf{9250} & \textbf{0.80} \\
        ~ & QwQ-32B & 13452 & 0.58 & 11436 & 0.72 & \underline{10528} & \underline{0.79} & \textbf{9981} & \textbf{0.83} \\
        ~ & Phi-4-Reasoning & 13732 & 0.50 & 11051 & 0.74 & \textbf{10624} & 0.73 & \underline{12603} & \textbf{0.77} \\
        ~ & RD-Qwen-1.5B & 10785 & 0.21 & 8895 & 0.47 & \underline{7125}  &  \underline{0.52} &  \textbf{7067} & \textbf{0.58} \\
        ~ & RD-Llama-3.1-8B & 13050 & 0.39 & \underline{8564} & \underline{0.70} & 9173 & 0.65 & \textbf{8199} & \textbf{0.73} \\
    \bottomrule
    \end{tabular}
    \label{tab:length_recali}
\end{table}

\subsubsection{Budget Constrained Accuracy (BCA)}
BCA aims to answer the following question: given a fixed token budget $B$, how does the accuracy shift accordingly? 
Intuitively, a smaller budget forces many correct but lengthy responses to terminate prematurely, thereby hurting the accuracy. 
Therefore, the BCA curve provides a clear depiction of reasoning efficiency.

\begin{figure}[ht]
    \includegraphics[width=\textwidth]{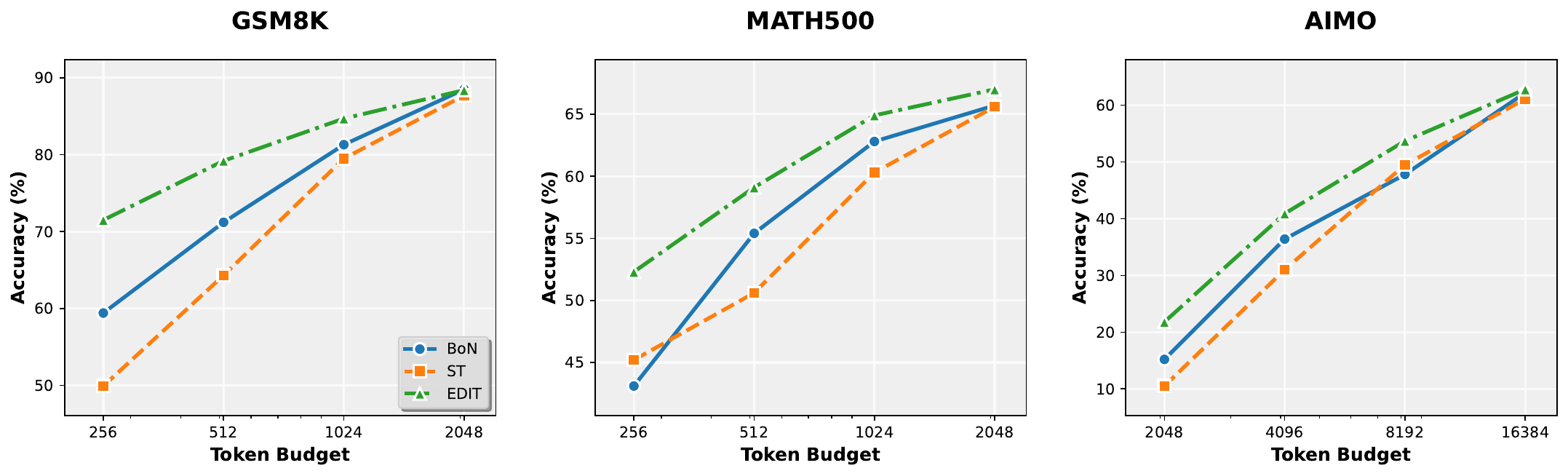}
    \caption{Accuracy variation under different token budget for three methods on three datasets. Accuracy values are averaged over 6 models on each dataset.}
    \label{fig:trend}
\end{figure}

To obtain the BCA curve, we collect many budget--accuracy datapoints from different methods.
As shown in~\Cref{fig:trend}, the accuracy generally improves as more generation budget is provided.
Notably, on all datasets, \methodname~leads by a large margin over the other two baselines.
This trend highlights the strong capability of~\methodname~to produce correct answers under limited computational budgets, and probably offers an explanation about why~\methodname~usually has the lowest calibrated length as in~\Cref{tab:length_recali}.

\subsection{Ablative Studies}
We conduct a series of experiments on ablative settings to verify the necessity of the proposed design components. 
In particular, we focus on three aspects: the choice of characteristic statistic, the maximum number of iterations, and the initial patience. 
Experiments are carried out on two models (R1-distilled-Qwen-1.5B and Llama-3.1-8B) under the corresponding ablative settings using the MATH dataset. The results are shown in~\Cref{tab:abl}.

There are some common patterns in the results of the two models. First, changing the length indicator causes performance degradation, which demonstrates the validity of employing a comprehensive metric that jointly considers multiple statistics. 

Next, regarding patience, a smaller patience value makes the search process overly cautious, preventing LRMs from producing concise solutions. In contrast, a larger patience value helps resist sampling noise and enables the search to continue toward the optimal solution.
Regarding the maximum number of iterations, the fixed sampling quota implies that increasing iterations reduces the sample size per iteration, resulting in answers with higher uncertainty. This effect is reflected in the observed drop in accuracy when the iteration count is set to 8.
Conversely, reducing the iteration upper bound has a similar effect to increasing patience: although more candidate reasoning paths are explored within each iteration, the search process may terminate prematurely at a shallow level, thereby failing to produce canonical answers.

\begin{table}[t]
    \centering
    \caption{MATH-500 test set Results on ablative configurations.}
    \begin{tabular}{l|ccccc}
    \toprule
       \multirow{2}{*}{Settings}  & \multicolumn{2}{c}{Llama-3.1} && \multicolumn{2}{c}{RD-Qwen-1.5B}  \\
       ~ & Acc & Length && Acc & Length \\
    \midrule
        Default ($\beta_0=1$, $\mathrm{max\_iter}=6$) & 52.4 & 108 && 62.4 & 249 \\
    \midrule
        Median only & 50.8 & 125 && 60.4 & 296 \\
        Minimum only & 52.2 & 139 && 61.0 & 303 \\
    \midrule
        $\beta_0=0$ & 51.6 & 157 && 60.6 & 281 \\
        $\beta_0=2$ & 49.2 & 128 && 59.2 & 252 \\
    \midrule
        $\mathrm{max\_iter}=4$ & 61.8 & 162 && 61.8 & 277 \\
        $\mathrm{max\_iter}=8$ & 58.5 & 129 && 57.2 & 316 \\
    \bottomrule
    \end{tabular}
    \label{tab:abl}
\end{table}

\section{Conclusion}
In this paper, we propose~\methodname~(\methodfullname), a test-time scaling method aimed at generating concise and correct reasoning paths.
\methodname~leverages distributional shifts  under different constraints 
discovered in our preliminary experiments, and designs a dual-goal search algorithm to reach the optimal tradeoff between simplicity and correctness. Experimental results across a broad range of models and datasets demonstrate the advantage and robustness compared to other strong baselines.
To mitigate the deficiency of using separate metrics to assess model overall performance, we further propose two alternative evaluation strategies for efficient reasoning, which can better calibrate the tradeoff between accuracy and length. 
We hope this work could provide new insights and inspiration for the LRM research community.
\section*{Limitations}
\label{sec:limit}
In this paper, we investigate the approach to search the shortest and correct solution on different reasoning tasks. 
We mainly focus on the statistical metrics on the same set of system configurations.
In fact, there are many possible directions from system's perspective that can further optimize the influence efficiency, such as parallel programming, better memory I/O, etc.
We believe future work can explore the intersection of these two areas to maximize the efficiency of large reasoning models.

\section*{Broader Impacts}
Besides what we mentioned in~\Cref{sec:limit}, we believe our work can inspire more researchers to explore more efficient reasoning models. While LRMs represent a promising future trend, they currently often suffer from low efficiency. There is an urgent need to reduce their potentially high computational cost.

\bibliography{custom}
\bibliographystyle{plain}

\clearpage
\appendix
\section{Experiment Details}
\subsection{Dataset Statistics}
We use three datasets for evaluation, all of which are in mathematical domain:
\begin{itemize}
    \item GSM8K~\cite{cobbe2021training}: The Grade School Math dataset contains 8.5K high quality linguistically diverse grade school math word problems, which is split into 7.5K training set and 1K test set. Problems in GSM8K require 2 to 8 steps to solve, and solutions primarily involve performing a sequence of elementary calculations using basic arithmetic operations (+ - / *) to reach the final answer. We use the test set for evaluation, which has 1,319 examples in total.
    \item MATH500~\cite{lightman2023let}: This dataset was released by OpenAI and initially used for reward-based training. We use a subset of the whole original dataset for evaluation, which comprises of 500 examples.
    \item AIMO~\cite{aime2025}: The AIMO (American Invitational Mathematics Olympics) dataset consists of 90 problems from AIME 2022-2024 competitions. This is the most challenging dataset among all the datasets.
\end{itemize}
\label{app:ds_stats}

\subsection{Model Specifications}
\label{app_model_spec}
There are seven targeted models being tested. All targeted models can be divided into three categories:
\begin{itemize}
    \item Instruction tuned model (Llama series): This category is the most typical instruction-following models. They only experience pretraining and supervised fine-tuning stage. 
    Additional red-teaming and human-preference alignment may be further applied to make its output fulfill the HHH (Helpful, Honest, and Harmless) principle.  
    Although demonstrating powerful general task solving capability, they show poor performance when encountering complex reasoning problems. 
    \item Reinforcement Learning (RL) enhanced reasoning model (QwQ-32B, Qwen3 series): These models are endowed with better reasoning capabilities through extra RL stage on large-scale reasoning data before traditional SFT and RL-based value-alignment stages.
    \item Distilled model (R1-distill-llama3-8B, R1-distill-Qwen-1.5B, Phi-4-reasoning-14B): These models are released along with DeepSeek-R1~\cite{guo2025deepseek}, which are fine-tuned on its automatically generated prompt-response pairs as illustrated in their papers.
\end{itemize}

\subsection{Evaluation Configurations}
We present all pivotal inference hyper-parameters in~\Cref{tab:inference_hp}. The inference scripts are implemented using the vLLM~\cite{kwon2023efficient}~library on four NVDIA A100 80G GPUs with tensor-level parallelism. We keep the same set of configurations for all tested models on the same dataset.

We put the core code at~\url{https://anonymous.4open.science/r/EDIT-F90B}.

\begin{table}[ht]
    \centering
    \begin{tabular}{c|ccc}
    \toprule
    Dataset   &  GSM8K & MATH & AIMO \\
    \midrule
    temperature    &  \multicolumn{3}{c}{0.7} \\
    top\_p         &  \multicolumn{3}{c}{0.9} \\
    num samples & \multicolumn{3}{c}{64} \\
    \midrule
    max token    & 2048 & 2048 & 16384 \\
    \bottomrule
    \end{tabular}
    \caption{Inference settings across three datasets.}
    \label{tab:inference_hp}
\end{table}

\label{app:eval_detail}

\section{More Statistical Results}

\subsection{Distribution of Length on Correct and Incorrect Answers}
\begin{figure}
    \centering
    \begin{subfigure}{0.95\linewidth}
        \includegraphics[width=\linewidth]{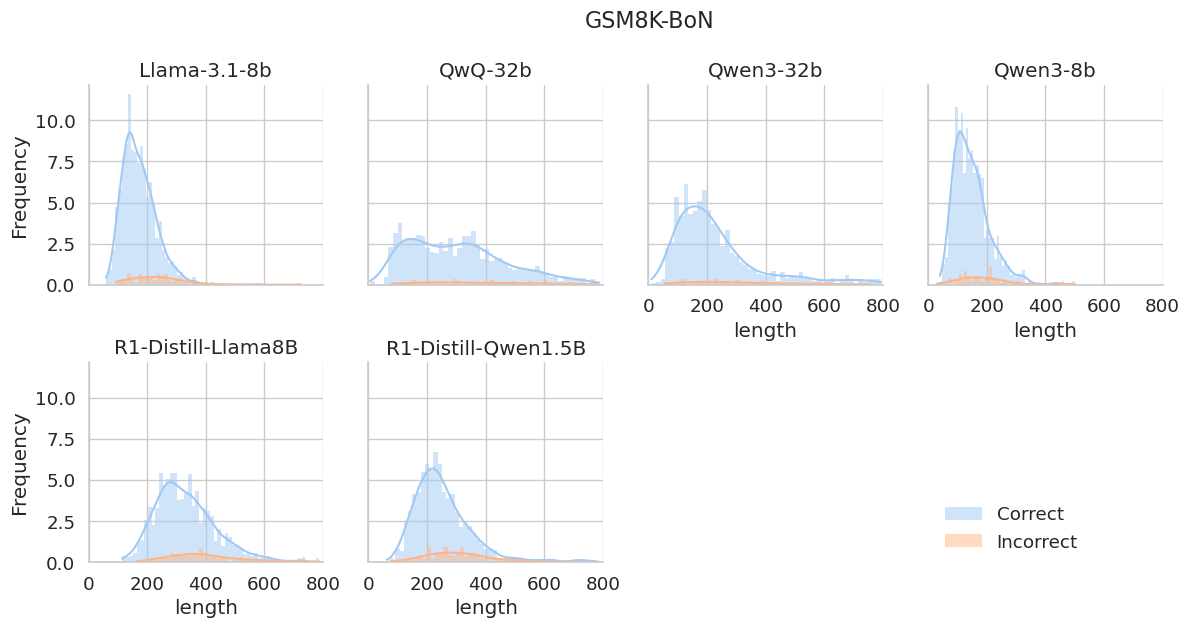}
        \caption*{}
    \end{subfigure}
    \vspace{-1em}
    \begin{subfigure}{0.97\linewidth}
        \includegraphics[width=\linewidth, trim=0 0 0 0]{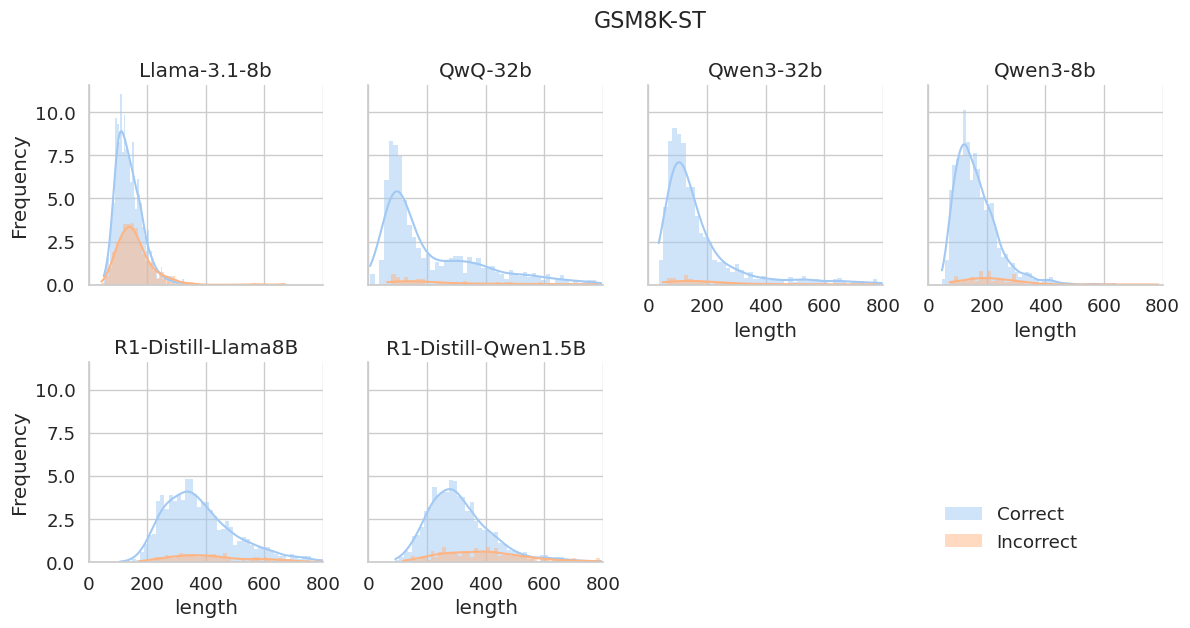}
        \caption*{}
    \end{subfigure}
    \vspace{-1em}
    \begin{subfigure}{0.95\linewidth}
        \includegraphics[width=\linewidth]{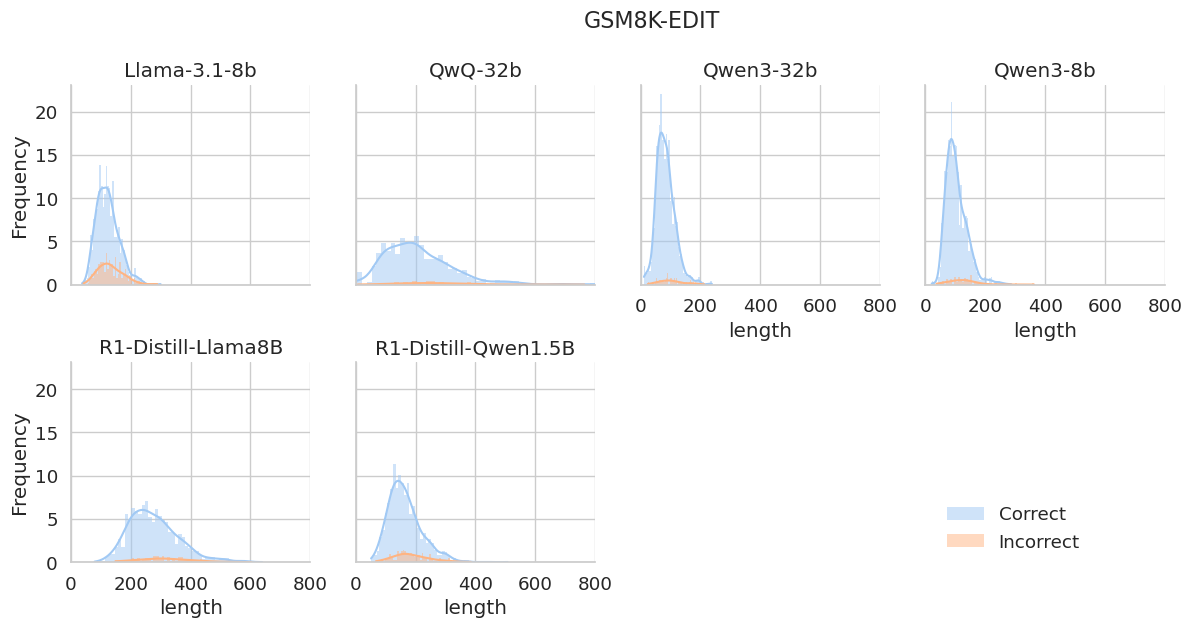}
        \caption*{}
    \end{subfigure}
    \vspace{-1em}
    \caption{Length distribution for correct and incorrect responses from three methods on GSM8K.}
    \label{fig:lendist_gsm8k}
\end{figure}

\begin{figure}
    \centering
    \begin{subfigure}{0.95\linewidth}
        \includegraphics[width=\linewidth]{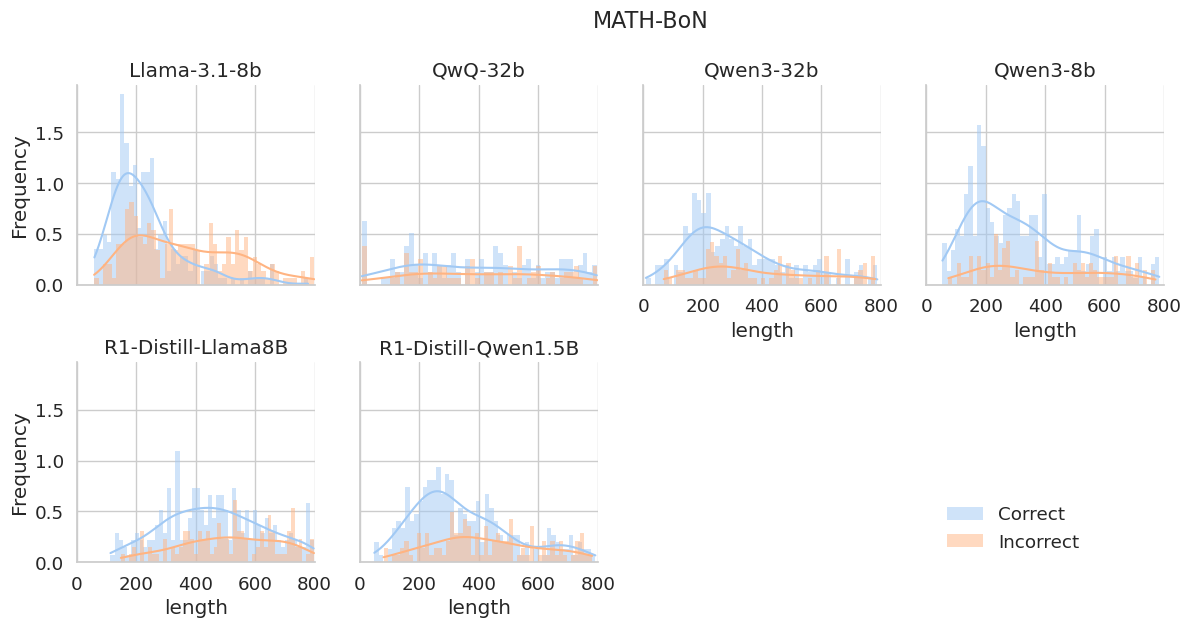}
        \caption*{}
    \end{subfigure}
    \vspace{-1em}
    \begin{subfigure}{0.95\linewidth}
        \includegraphics[width=\linewidth, trim=0 0 0 0]{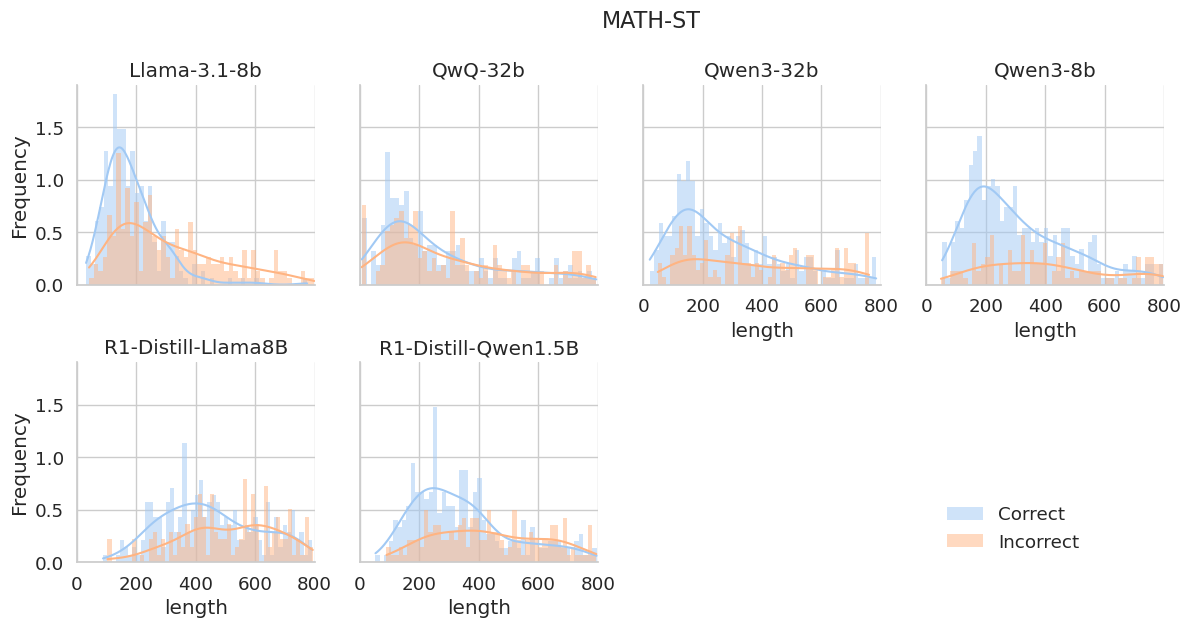}
        \caption*{}
    \end{subfigure}
    \vspace{-1em}
    \begin{subfigure}{0.95\linewidth}
        \includegraphics[width=\linewidth]{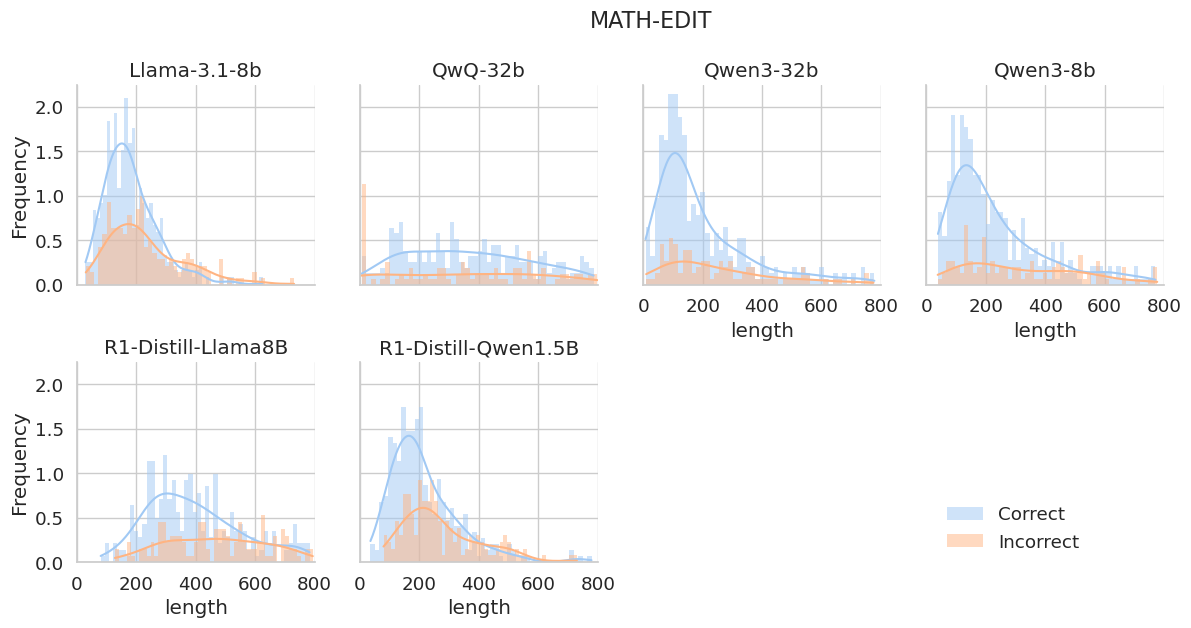}
        \caption*{}
    \end{subfigure}
    \vspace{-1em}
    \caption{Length distribution for correct and incorrect responses from three methods on MATH.}
    \label{fig:lendist_math}
\end{figure}

We further plot the length distribution of correct/incorrect answers on GSM8K test set in~\Cref{fig:lendist_gsm8k}~and~\Cref{fig:lendist_math}. It can be found that for all models, compared to other two baselines, the distribution of~\methodname~shifts to the left---indicating there the huge reduction in lengths. Besides, the area ratio between the blue and orange bars also increases sharply, which means~\methodname~gets more correct answers.

\subsection{Generation Examples}
We show some typical examples of responses from these three methods, as shown in~\Cref{tab:example1,tab:example2}. In the simple GSM8K example,~\methodname~reaches the correct answer using much less tokens than the other methods. 
In the hard MATH500 example, where the model has to understand the code in the question, BoN fails to get the correct answer due to the insufficient uncertainty of the correct answer. In contrast,~\methodname~successfully yields the correct answer.

\begin{longtable}{l}
    \caption{An GSM8K problem solved by~\methodname~and several baselines and R1-Distill-Qwen1.5B. Green titled boxes represent that the solution is both correct and the shortest. We remove some special tokens for better readability. Blue titled boxes mean that only answer is correct. Red titled boxes fail to produce the correct answer. } 
    \label{tab:example1} \\
    
    \begin{tcolorbox}[
      colback=gray!10,        
      coltitle=white,         
      title=GSM8K Question,     
      colbacktitle=black!70,     
      fonttitle=\bfseries
      ]
        There are some oranges in a basket. Ana spends 3 minutes peeling an orange and Jane spends 4 minutes doing the same. If Ana and Jane start picking oranges from this basket to peel at the same time, how many more oranges will Ana have peeled than Jane after an hour? \\
        Answer: 5
    \end{tcolorbox} \\
    \begin{tcolorbox}[
      colback=gray!10,        
      coltitle=black,         
      title=\methodname~(142 Tokens),     
      colbacktitle=green!60,     
      fonttitle=\bfseries
      ]
    Step 1: Determine the peeling rates of Ana and Jane. \\
        - Ana peels 1 orange per 3 minutes. \\
        - Jane peels 1 orange per 4 minutes. \\
        
    Step 2: Calculate the number of oranges each can peel in an hour (60 minutes). \\
        - Ana: \( \frac{60}{3} = 20 \) oranges. \\
        - Jane: \( \frac{60}{4} = 15 \) oranges. \\
    Step 3: Find the difference in oranges peeled. \\
        - Ana peels \( 20 - 15 = 5 \) more oranges than Jane. \\

\textbf{Answer}: \boxed{5}
    \end{tcolorbox} \\
    
    \begin{tcolorbox}[
      colback=gray!10,        
      coltitle=black,         
      title=BoN~(226 Tokens),     
      colbacktitle=blue!20,     
      fonttitle=\bfseries
      ]
    First, Let’s define, Ana’s rate = 1 orange per 3 minutes, Jane’s rate = 1 orange per 4 minutes. In 1 hour (60 minutes), Ana peels 60 ÷ 3 = 20 oranges. Jane peels 60 ÷ 4 = 15 oranges. Therefore, Ana peeled 5 more oranges than Jane. \\

    Step-by-Step Explanation: \\

    1. **Determine Individual Rates:**  \\
       - Ana peels 1 orange in 3 minutes.   \\
       - Jane peels 1 orange in 4 minutes.  \\
    
    2. **Calculate Oranges Plucked in an Hour:**    \\
       - One hour = 60 minutes.                     \\
       - **Ana:** \( 60 \div 3 = 20 \) oranges.      \\
       - **Jane:** \( 60 \div 4 = 15 \) oranges.    \\
    
    3. **Find the Difference:**
       - \( 20 - 15 = 5 \) oranges.
    
    Final Answer: \boxed{5}
    \end{tcolorbox} \\

    \begin{tcolorbox}[
      breakable,
      colback=gray!10,        
      coltitle=black,         
      title=ST~(278 Tokens),     
      colbacktitle=blue!20,     
      fonttitle=\bfseries
      ]
    First, Let’s define, Ana’s rate = 1 orange per 3 minutes, Jane’s rate = 1 orange per 4 minutes. In 1 hour (60 minutes), Ana peels 60 ÷ 3 = 20 oranges. Jane peels 60 ÷ 4 = 15 oranges. Therefore, Ana peeled 5 more oranges than Jane. \\
    
    Step-by-Step Explanation: \\
    
    1. **Determine Individual Rates:** \\
       - Ana peels 1 orange in 3 minutes. \\ 
       - Jane peels 1 orange in 4 minutes. \\
    
    2. **Calculate Oranges Plucked in an Hour:** \\
       - One hour = 60 minutes. \\
       - **Ana:** \( 60 \div 3 = 20 \) oranges. \\
       - **Jane:** \( 60 \div 4 = 15 \) oranges. \\
    
    3. **Find the Difference:** \\
       - \( 20 - 15 = 5 \) oranges. \\
    
    **Final Answer:** \boxed{5}
    \end{tcolorbox}
\end{longtable}

\newpage
\begin{longtable}{c}
    \caption{A MATH500 problem solved by~\methodname~and several baselines and R1-distill-Qwen-1.5B.} \\
    \label{tab:example2}
    
    \begin{tcolorbox}[
      colback=gray!10,        
      coltitle=white,         
      title=MATH500 Question,     
      colbacktitle=black!70,     
      fonttitle=\bfseries
      ]
        Let $z = 2 + \sqrt{2} - (3 + 3 \sqrt{2})i$, and let $c = 2 - 3i$.  Let $w$ be the result when $z$ is rotated around $c$ by $\frac{\pi}{4}$ counter-clockwise.

[asy] \\
unitsize(0.6 cm);

pair C, W, Z; \\

$Z = (2 + \sqrt(2), -3 - 3*\sqrt(2))$;
$C = (2,-3)$;
$W = \mathrm{rotate}(45,C)*(Z)$;

draw(Z--C--W);

dot("$c$", C, N); \\
dot("$w$", W, SE); \\
dot("$z$", Z, S); \\
label("$\frac{\pi}{4}$", C + (0.6,-1));

[/asy]
\\
Find $w.$

\textbf{Answer}: $6 - 5i$
    \end{tcolorbox} \\
    \begin{tcolorbox}[
      colback=gray!10,        
      coltitle=black,         
      title=\methodname~(667 Tokens),     
      colbacktitle=green!60,     
      fonttitle=\bfseries
      ]

Step 1: Identify the given points and transformations.

We are given:
- \( z = 2 + \sqrt{2} - (3 + 3\sqrt{2})i \)
- \( c = 2 - 3i \)
- \( w \) is the result of rotating \( z \) around \( c \) by \( \frac{\pi}{4} \) counter-clockwise.

Step 2: Translate the complex number \( z \) so that \( c \) is at the origin.

Translation is done by subtracting \( c \) from \( z \):
\[
z' = z - c = (2 + \sqrt{2} - (3 + 3\sqrt{2})i) - (2 - 3i) = (\sqrt{2}) - (3\sqrt{2})i
\]

Step 3: Rotate the translated complex number \( z' \) by \( \frac{\pi}{4} \).

The rotation formula is:
\[
w' = z' \cdot \left(\cos\left(\frac{\pi}{4}\right) + i\sin\left(\frac{\pi}{4}\right)\right) = z' \cdot \left(\frac{\sqrt{2}}{2} + i\frac{\sqrt{2}}{2}\right)
\]
Multiply \( z' \) by this rotation factor:
\[
w' = (\sqrt{2} - 3\sqrt{2}i) \cdot \left(\frac{\sqrt{2}}{2} + i\frac{\sqrt{2}}{2}\right)
\]

Step 4: Perform the multiplication.

Multiply the terms:
\[
w' = \sqrt{2} \cdot \frac{\sqrt{2}}{2} + \sqrt{2} \cdot i\frac{\sqrt{2}}{2} - 3\sqrt{2}i \cdot \frac{\sqrt{2}}{2} - 3\sqrt{2}i \cdot i\frac{\sqrt{2}}{2}
\]
Simplify each term:
- \( \sqrt{2} \cdot \frac{\sqrt{2}}{2} = 1 \)
- \( \sqrt{2} \cdot i\frac{\sqrt{2}}{2} = i1 \)
- \( -3\sqrt{2}i \cdot \frac{\sqrt{2}}{2} = -3i1 \)
- \( -3\sqrt{2}i \cdot i\frac{\sqrt{2}}{2} = 3 \) (since \( i^2 = -1 \))
\[
w' = 1 + i - 3i + 3 = 4 - 2i
\]

Step 5: Translate back by adding \( c \).

Add \( c \) to \( w' \):
\[
w = w' + c = (4 - 2i) + (2 - 3i) = 6 - 5i
\]

Answer:
\boxed{6 - 5i}

\end{tcolorbox} \\

\begin{tcolorbox}[
  breakable,
  colback=gray!10,        
  coltitle=black,         
  title=BoN~(610 Tokens but wrong answer),     
  colbacktitle=red!40,     
  fonttitle=\bfseries
  ]
Let's find the value of \( w \) after rotating \( z \) around \( c \) by \( \frac{\pi}{4} \) counter-clockwise.

**Step 1: Identify the given points and complex numbers**

- \( z = 2 + \sqrt{2} - (3 + 3 \sqrt{2})i \)
- \( c = 2 - 3i \)

**Step 2: Translate \( z \) so that the center \( c \) becomes the origin**

To rotate around \( c \), we translate \( z \) by subtracting \( c \):

\[
z' = z - c = (2 + \sqrt{2} - 2) + (-3 - 3\sqrt{2} - (-3))i = \sqrt{2} - 3\sqrt{2}i
\]

**Step 3: Apply the rotation by \( \frac{\pi}{4} \)**

The rotation formula for a complex number \( z' \) by \( \theta \) radians is:

\[
w = z' \cdot e^{i\theta} = z' \cdot (\cos\theta + i\sin\theta)
\]

For \( \theta = \frac{\pi}{4} \):

\[
e^{i\frac{\pi}{4}} = \cos\frac{\pi}{4} + i\sin\frac{\pi}{4} = \frac{\sqrt{2}}{2} + i\frac{\sqrt{2}}{2}
\]

Multiply \( z' \) by this:

\[
w = (\sqrt{2} - 3\sqrt{2}i)\left(\frac{\sqrt{2}}{2} + i\frac{\sqrt{2}}{2}\right)
\]

**Step 4: Perform the multiplication**

Multiply each term:

\[
w = \sqrt{2} \cdot \frac{\sqrt{2}}{2} + \sqrt{2} \cdot i\frac{\sqrt{2}}{2} - 3\sqrt{2}i \cdot \frac{\sqrt{2}}{2} - 3\sqrt{2}i \cdot i\frac{\sqrt{2}}{2}
\]

Simplify each term:

\[
w = \frac{2}{2} + i\frac{2}{2} - i\frac{6}{2} - i^2\frac{6}{2}
\]

\[
w = 1 + i - 3i - (-1)\cdot3
\]

\[
w = 1 - 2i + 3 = 4 - 2i
\]

Final Answer: \boxed{4 - 2i}

    \end{tcolorbox} \\

    \begin{tcolorbox}[
      colback=gray!10,        
      coltitle=black,         
      title=ST~(705 Tokens),     
      colbacktitle=blue!20,     
      fonttitle=\bfseries
      ]
Step 1: Identify the given points and transformations.

We are given:
- \( z = 2 + \sqrt{2} - (3 + 3\sqrt{2})i \)
- \( c = 2 - 3i \)
- \( w \) is the result of rotating \( z \) around \( c \) by \( \frac{\pi}{4} \) counter-clockwise.

Step 2: Translate the complex number \( z \) so that \( c \) is at the origin.

Translation is done by subtracting \( c \) from \( z \):
\[
z' = z - c = (2 + \sqrt{2} - (3 + 3\sqrt{2})i) - (2 - 3i) = (\sqrt{2}) - (3\sqrt{2})i
\]

Step 3: Rotate the translated complex number \( z' \) by \( \frac{\pi}{4} \).

The rotation formula is:
\[
w' = z' \cdot \left(\cos\left(\frac{\pi}{4}\right) + i\sin\left(\frac{\pi}{4}\right)\right) = z' \cdot \left(\frac{\sqrt{2}}{2} + i\frac{\sqrt{2}}{2}\right)
\]
Multiply \( z' \) by this rotation factor:
\[
w' = (\sqrt{2} - 3\sqrt{2}i) \cdot \left(\frac{\sqrt{2}}{2} + i\frac{\sqrt{2}}{2}\right)
\]

Step 4: Perform the multiplication.

Multiply the terms:
\[
w' = \sqrt{2} \cdot \frac{\sqrt{2}}{2} + \sqrt{2} \cdot i\frac{\sqrt{2}}{2} - 3\sqrt{2}i \cdot \frac{\sqrt{2}}{2} - 3\sqrt{2}i \cdot i\frac{\sqrt{2}}{2}
\]
Simplify each term:
- \( \sqrt{2} \cdot \frac{\sqrt{2}}{2} = 1 \)
- \( \sqrt{2} \cdot i\frac{\sqrt{2}}{2} = i1 \)
- \( -3\sqrt{2}i \cdot \frac{\sqrt{2}}{2} = -3i1 \)
- \( -3\sqrt{2}i \cdot i\frac{\sqrt{2}}{2} = 3 \) (since \( i^2 = -1 \))
\[
w' = 1 + i - 3i + 3 = 4 - 2i
\]

Step 5: Translate back by adding \( c \).

Add \( c \) to \( w' \):
\[
w = w' + c = (4 - 2i) + (2 - 3i) = 6 - 5i
\]
Final Answer: \boxed{6 - 5i}
    \end{tcolorbox}
\end{longtable}


\newpage

\end{document}